\documentclass{article}
\usepackage{spconf,amsmath,epsfig}
\usepackage[colorlinks=false, pdfborder={0 0 0}]{hyperref}
\usepackage{multirow}
\usepackage{colortbl}
\usepackage{tikz}
\usepackage{caption}
\usepackage{graphicx}
\usepackage{subcaption}

\title{Predicting Crop Yield With Machine Learning: An Extensive Analysis Of Input Modalities And Models On a Field and sub-field Level}
\name{
\begin{tabular}{@{}c@{}}
Deepak Pathak$^{*, 1, 2}$ \qquad Miro Miranda$^{*,1, 2}$ \qquad Francisco Mena$^{1, 2}$ \qquad Cristhian Sanchez$^{1, 2}$ \\
Patrick Helber$^{3}$ \qquad Benjamin Bischke$^{3}$ \qquad Peter Habelitz$^{3}$ \qquad Hiba Najjar$^{1, 2}$ \\
Jayanth Siddamsetty$^{2}$ \qquad Diego Arenas$^{2}$ \qquad Michaela Vollmer$^{2}$ \qquad Marcela Charfuelan$^{2}$ \\
Marlon Nuske$^{2}$ \qquad Andreas Dengel$^{1, 2}$
\end{tabular}
\thanks{*both authors contributed equally to this work}
}
 \address{$^{1,}$University of Kaiserslautern-Landau (RPTU), Kaiserslautern, Germany\\
     $^{2}$German Research Center for Artificial Intelligence (DFKI), Kaiserslautern, Germany\\
     $^{3}$Vision Impulse GmbH, Kaiserslautern, Germany }

\begin{document}
%
\maketitle
\begin{abstract}

We introduce a simple yet effective early fusion method for crop yield prediction that handles multiple input modalities with different temporal and spatial resolutions. We use high-resolution crop yield maps as ground truth data to train crop and machine learning model agnostic methods at the sub-field level. We use Sentinel-2 satellite imagery as the primary modality for input data with other complementary modalities, including weather, soil, and DEM data. The proposed method uses input modalities available with global coverage, making the framework globally scalable. We explicitly highlight the importance of input modalities for crop yield prediction and emphasize that the best-performing combination of input modalities depends on region, crop, and chosen model.
\end{abstract}
\section{Introduction}
\label{sec:intro}

Yield prediction is an essential task in the agricultural sector. Yet, it is still challenging due to multidimensional factors defining the yield, including environmental factors, management, the genotype, and their interactions. Providing accurate yield prediction not only supports industry and farmers in decision-making such as pest control, fertilization, and harvest time prediction, but also policymakers. In light of changing and fluctuating climate conditions, reliable yield predictions are increasingly challenging and are nowadays addressed from multiple perspectives. Here, machine learning has played an increasingly important role in recent years \cite{van2020crop}. 
With the rise of remote sensing technology, yield prediction can be addressed from a large-scale perspective, as data is available globally with high temporal frequencies. This offers various opportunities for crop monitoring and particularly for yield prediction \cite{hunt2019high,kayad2019monitoring}.
Currently, models are trained on diverse sets of remotely sensed input modalities such as satellite imagery, weather, soil, and Digital Elevation Model (DEM) data. \cite{moravec2017digital, kayad2019monitoring, schwalbert2020satellite}. Although it is known that all the mentioned modalities are good yield predictors, only a subset is included in most studies. It still needs to be determined if including multiple modalities is beneficial for model performance. 
In addition, most studies focus on a narrow regional level with single crop cultivars and few training years, making models highly susceptible to regional and temporal overfitting. It is, moreover, still an open question if machine learning can predict crop yields consistently over years, regions, and crop types.

In this research, we present an operational approach to multimodal yield prediction at the pixel level in $10m$ resolution, referred to as sub-field, that is crop and region independent and globally scalable. We analyze the importance of input modalities, such as satellite imagery, and additional modalities, including weather, soil, and DEM data. A simple but effective way of data fusion is proposed to combine data with different temporal and spatial resolutions. Results are evaluated on a large dataset containing different countries, crops, and years at field and sub-field level.

\begin{figure*}[t]
\begin{subfigure}{.5\textwidth}
  \centering
  \includegraphics[width=\linewidth]{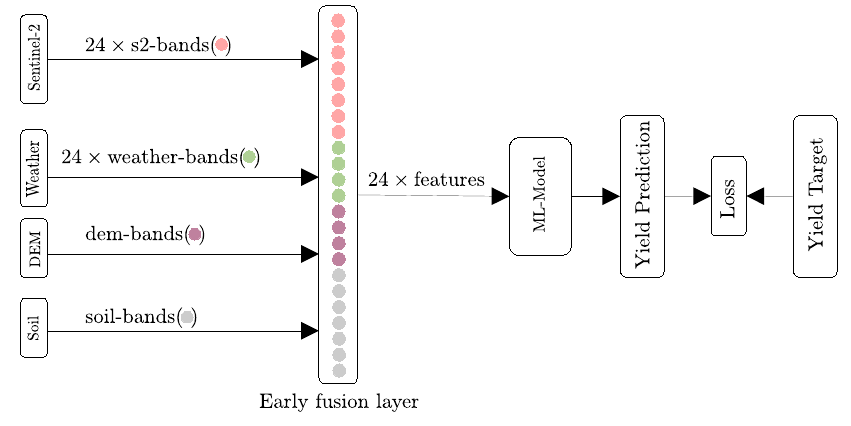}
  \caption{Early Fusion Model}
  \label{fig:early_fusion_model}
\end{subfigure}%
\begin{subfigure}{.5\textwidth}
  \centering
  \includegraphics[width=\linewidth]{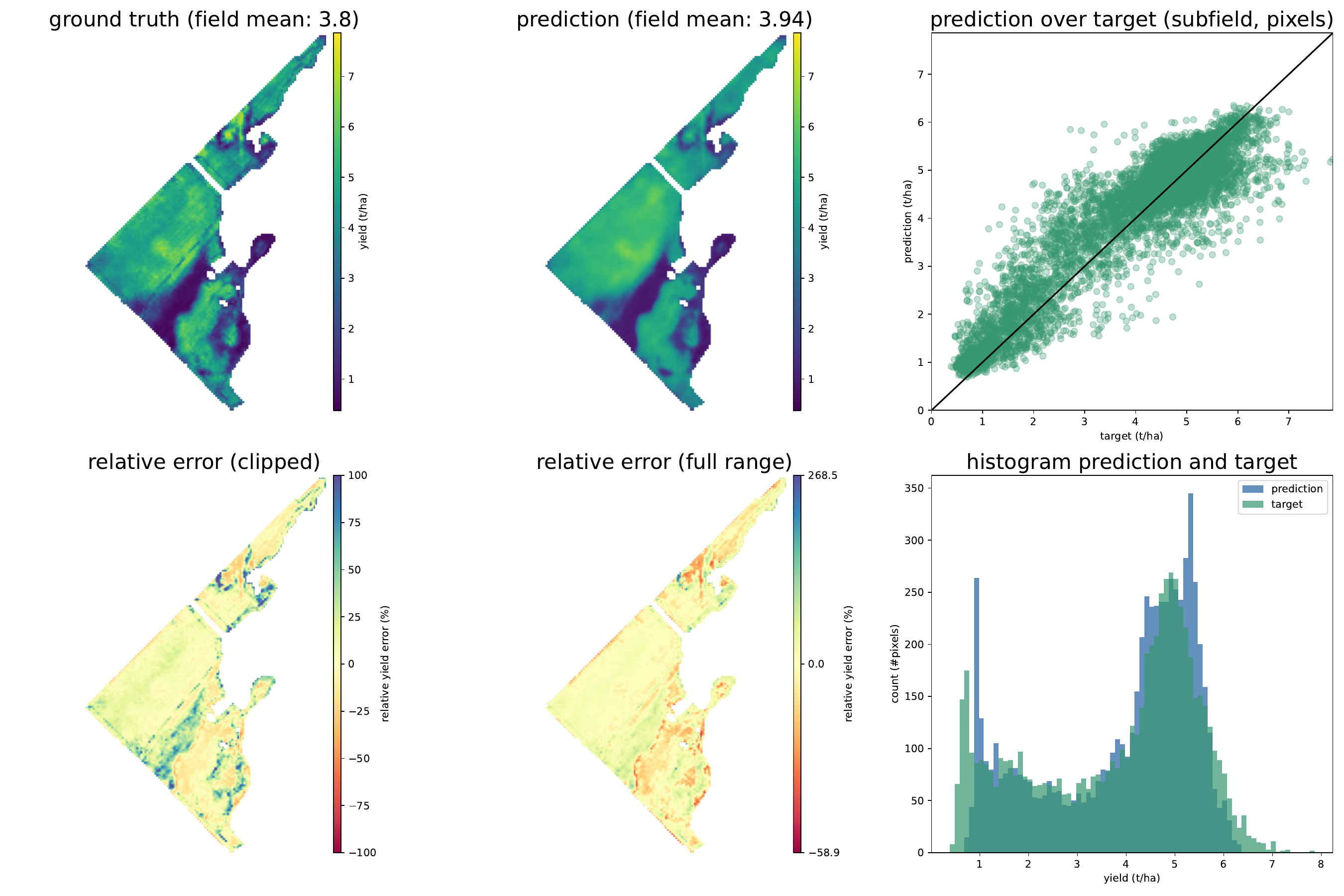}
  \caption{Prediction, target and yield distribution.}
  \label{fig:field_evaluation}
\end{subfigure}
\caption{(a) Framework for multimodal data fusion for yield predictions. Multiple modalities with different spatial and temporal resolutions are fused at the input level. A machine learning model is then trained pixel-wise to produce yield predictions in $10m$ resolution. (b) Performance plots for visual inspection of a single field. Yield data from soybean in Argentina is shown, harvested in 2021. The model was trained on Sentinel-2 and DEM data. Upper left: ground truth yield map, upper middle: pixel-based yield prediction, upper right: scatterplot comparing predictions with ground truth data, lower left: relative prediction clipped at 100\%, lower middle: relative prediction error in full range, lower right: distribution plot of predictions against the target.}
\label{fig:main_fig}
\end{figure*}

\section{Material \& Methods}
We include data over different countries, crop types and years. In detail, we use data coming from \textit{ Germany, Argentina, and Uruguay}. For each country, different crop types are available, including \textit{wheat, rapeseed, and soybean}. For yield forecasting, we use gradient boosting and deep learning-based methods.
\subsection{Data}
For training, yield data is used as ground truth, combined with remotely sensed input data available with global coverage as predictive features.

\paragraph*{Yield Data}
Yield data from combine harvesters on a sub-field level is used as ground truth data. While harvesting, the combine harvester with yield monitors drives through the field, collecting equidistant data points in high spatial resolution. Each data point is characterized by different features such as the geographic coordinate, the amount of yield in t/ha, the yield moisture in $\%$. We use a standardized data pre-processing pipeline to harmonize the raw yield data. This includes reprojecting the coordinate reference system, standardization of feature naming, and removing erroneous values for position, timestamp, yield, moisture, and non-activated harvesters. Zero yield points and biologically infeasible points are removed. In addition, data points are filtered by statistical thresholds, meaning that a yield point must be within three standard deviations. For more details, we refer the reader to \cite{sanchez2023}. The resulting point vector data is rasterized into $10m$ resolution yield maps aligning with satellite imagery raster data. Tab. \ref{table:datasets} gives an overview of the used yield datasets.

\paragraph*{Sentinel-2 Data}
All experiments use cloud-free Sentinel-2 (L2A) images (S2) with $10m$ resolution for model training. Spectral bands with lower resolutions are upsampled to $10m$ resolution, resulting in twelve spectral bands. Images are collected within the growing period, i.e., between each field's seeding and harvesting date.
\paragraph*{Additional Data Modalities}
In addition to satellite imagery, we select a set of data modalities that are known to play a role in plant development and yield formation.
We can categorize Additional Data Modalities (ADM) into weather data, soil data, and DEM data. Weather data for each field is derived from the ECMWF Reanalysis (ERA5) \cite{hersbach2020era5}, soil data from SoilGrids in $250m$ resolution \cite{10.1371/journal.pone.0169748}, and DEM data from NASA’s Shuttle Radar Topography Mission (SRTM)\cite{farr2000shuttle} in $30m$ resolution. We prepared the ADM based on the bounds of the field using ground truth data. For soil and DEM data, raster images are created and upsampled to $10m$ resolution using a cubic spline interpolation. For soil, we use all eight available soil properties, i.e. \textit{cec, cfvo, nitrogen, phh2o, sand, silt, soc, clay} at depth of 0-5, 5-15, and 15-30 cm. For DEM, we used the RichDEM \cite{RichDEM} tool for feature engineering and deriving more features that include \textit{aspect, curvature, dem, slope, twi}. Weather data is aggregated for each day at field level for minimum, maximum, and mean temperature and total precipitation.   

\subsection{Data Preprocessing}
\label{sec:preprocessing}
For each field, input data is represented as a sequence of 24 timesteps defining two calendar years (a sample for each month) with the harvesting date in the second year. We mask all samples outside the crop season, i.e., before seeding and after harvest. S2 images are used as reference data to create 24 timesteps \cite{helber2023} by selecting the best cloud free S2 image among all images within each time interval, and features from other modalities are concatenated for each timestep with S2 features. Daily weather data is aggregated by summing all values between each time interval based on the dates of the selected S2 images. Soil and DEM features are vectorized and repeated at each timestep. This preprocessing results in a multivariate time series in which each sample represents a raster pixel of the yield map, with a maximum of 45 features at each timestep, depending on the selected ADM. 

\begin{table}[!hbt]
    \centering
    \caption{\label{table:datasets}
        Yield map (fields) data per country and crop type for different years.
        }
    \resizebox{\columnwidth}{!}{

\begin{tabular}{|cc|cccc|} 
\hline
\multicolumn{1}{|c|}{\textbf{Country}} & \textbf{Years} & \multicolumn{1}{c|}{\textbf{Rapeseed}} & \multicolumn{1}{c|}{\textbf{Wheat}} & \multicolumn{1}{c|}{\textbf{Soybean}} & \textbf{Sum}  \\ 
\hline
\textbf{Germany}                       & 2016-2022      & 111                                    & 188                                 & 0                                     & 299           \\
\textbf{Uruguay}                       & 2018-2021      & 0                                      & 0                                   & 486                                   & 486           \\
\textbf{Argentina}                     & 2017-2022      & 0                                      & 0                                   & 192                                   & 192           \\ 
\hline
\multicolumn{2}{|c|}{\textbf{Sum}}                      & 111                                    & 188                                 & 678                                   & \textbf{977}  \\
\hline
\end{tabular}
    } 
\end{table}

\subsection{Methods}
We used state-of-the-art machine learning and deep learning models to capture in-field variability, namely \textit{Light Gradient-Boosting Machine (LGBM)} \cite{ke_2017_lightGBM} and \textit{Long Short-Term Memory (LSTM)} \cite{hochreiter_1997_lstm}. An overview of the proposed framework is illustrated in Fig.\ref{fig:early_fusion_model}. Following the early fusion method \cite{pradeep_2010_mml_fusion}, a multivariate time series is created, wherein each timestep represents concatenated features described in sec. \ref{sec:preprocessing}. The time series is further fed to a machine learning model for a regression task, where each sample represents a pixel with $10m$ resolution based on S2 images. 
For LGBM, the preprocessed data is vectorized by concatenating all timesteps into one vector. For LSTM, the preprocessed data is used sequentially, one timestep at a time to feed the model.
The LGBM model is trained using regression objective with \textit{gbdt} boosting type, learning rate as 0.1, and early stopping round as 10.
In the LSTM model, 2 stacked LSTM layers with 128 hidden units are used, followed by two fully-connected layers with 128 and 1 neurons respectively, separated with a ReLU non-linear activation and batch-normalization to output the predicted yield value. The LSTM model is trained using ADAM optimizer with a fixed learning rate of 0.001 and batch size 1024 for 50 epochs. An early stopping method is used to halt training if the model does not improve for 8 consecutive epochs on the validation data. To avoid overfitting, stratified grouped K-fold cross validation is used, grouped with field name and stratified with farm name. Here, a farm represents either a set of fields operated by a farmer or geographically nearby fields in case farmer information is unavailable. We report scores as the average over K-Folds, using 10-Folds in all experiments. 

\section{Results \& Evaluation}
We quantitatively and qualitatively evaluate model performance. For quantitative evaluation at field and sub-field level regression, 
we use the Mean Absolute Percentage Error (MAPE) and
the coefficient of determination, \textit{R-squared}($R^2$). For qualitative evaluation, a three point-guideline is used. (1) In-field variability: the model should capture sub-field differences, (2) low prediction error: low pixel-wise prediction error, (3) distribution match: prediction and target distribution must be close to each other. We consider a model trained on S2 data only as a baseline and investigate the contribution of ADM. 
Tab. \ref{table:mx_results} shows the effect of including ADM in addition to S2 on the performance of the LSTM model in Argentina for soybean.
We observe that although all ADM improve performance, DEM data in addition to S2 boosts the performance most, i.e. an $R^2$ of $0.82$, resulting in an improvement of 8 percentage points (p.p.) over S2 only. Similarly, in \textit{Germany}, for rapeseed, we observe an $R^2$ of $0.78$ by using S2 and soil data, meaning an improvement of 13 p.p. over S2 data only. Moreover, in Tab. \ref{table:results}, similar experiments are done for all other crops and regions. We present results of the best performing combination of model and different modalities in the context of field and sub-field level performances. Looking at the qualitative evaluation, we see reasonable performances of all models over countries, crops, and years. We note that the presented framework captures in-field variability. In addition, we observe low prediction errors and good distribution match in numerous instances. An example is shown in Fig.\ref{fig:field_evaluation}.
\begin{table}[!bth]
    \tiny
    \centering
    \caption{\label{table:mx_results}
        Contribution of different modalities in soybean yield prediction for Argentina using the LSTM model.
        }
    \resizebox{\columnwidth}{!}{


\begin{tabular}{|c|cccc|} 
\hline
\multirow{2}{*}{\textbf{Modalities}}                  & \multicolumn{2}{c|}{\textbf{FIELD}}              & \multicolumn{2}{c|}{\textbf{SUBFIELD}}  \\ 
\cline{2-5}
                                                      & \textbf{MAPE} & \multicolumn{1}{c|}{\textbf{R2}} & \textbf{MAPE} & \textbf{R2}             \\ 
\hline
\rowcolor[rgb]{0.95,0.95,0.95} S2-Weather-Soil-DEM & 0.11          & 0.76                             & 0.24          & 0.63                    \\
\textbf{S2-DEM}                                       & \textbf{0.09} & \textbf{0.82}                    & \textbf{0.24} & \textbf{0.65}           \\
\rowcolor[rgb]{0.95,0.95,0.95} S2-Soil             & 0.1           & 0.76                             & 0.25          & 0.61                    \\
S2-Weather                                            & 0.11          & 0.78                             & 0.25          & 0.63                    \\
\rowcolor[rgb]{0.95,0.95,0.95} S2                  & 0.11          & 0.74                             & 0.25          & 0.61                    \\
\hline
\end{tabular}
    }
\end{table}

\begin{table}[!bth]
    \centering
    \caption{\label{table:results}
        Results show the best-performing combination of different modalities and ML methods for distinct crops and countries at the field and sub-field level crop yield prediction.
        }
    \resizebox{\columnwidth}{!}{

\begin{tabular}{|c|c|c|c|cccc|} 
\hline
\multicolumn{4}{|c|}{\textbf{Evaluation}}                                                       & \multicolumn{2}{c|}{\textbf{Field}}                                & \multicolumn{2}{c|}{\textbf{Sub-field}}        \\ 
\hline
\textbf{Model}                         & \textbf{Modalities} & \textbf{Crop} & \textbf{Country} & \textit{\textbf{MAPE}} & \multicolumn{1}{c|}{\textit{\textbf{R2}}} & \textit{\textbf{MAPE}} & \textit{\textbf{R2}}  \\ 
\hline
\rowcolor[rgb]{0.949,0.949,0.949} LSTM & S2-DEM              & Soybean       & Argentina        & 0.09                   & 0.82                                      & 0.24                   & 0.65                  \\
LSTM                                   & S2-Soil             & Rapeseed      & Germany          & 0.15                   & 0.78                                      & 0.39                   & 0.45                  \\
\rowcolor[rgb]{0.949,0.949,0.949} LGBM & S2-Weather-Soil-DEM & Soybean       & Uruguay          & 0.2                    & 0.77                                      & 1.02                   & 0.42                  \\
LGBM                                   & S2-Weather-Soil-DEM & Wheat         & Germany          & 0.09                   & 0.68                                      & 0.29                   & 0.37                  \\
\hline
\end{tabular}
    }
\end{table}
\section{Conclusion \& Outlook}
State-of-the-art machine learning models are well suited for yield predictions over countries, crops, and years. Surprisingly, we observe regional different feature importance, resulting in the selection of input features being essential for ML-based crop yield prediction. Models trained on multimodal data outperform models trained on satellite imagery only. Adding additional modalities with low spatial resolution significantly increases field-level performance and, moreover, improves sub-field level performance. In this study, we focused on evaluating early fusion methods. Nevertheless, it is still unclear whether other fusion methods can better extract yield-driving features and thus learn to avoid insignificant modalities for the crop-region combination. Also, it is still to be examined if more data modalities, including expert knowledge, would further contribute to the models' performance. 

\section{Acknowledgement}
The research results presented are part of a large collaborative project on agricultural yield predictions, which was partly funded through the ESA InCubed Programme (\url{https://incubed.esa.int/}) as part of the project AI4EO Solution Factory (\url{https://www.ai4eo-solution-factory.de/}). H.N. and F.M. acknowledge support through a scholarship of the University of Kaiserslautern-Landau.


\bibliographystyle{IEEEbib}
\bibliography{refs}

\end{document}